\documentclass[letterpaper, 10 pt, conference]{ieeeconf}  
\IEEEoverridecommandlockouts                            

\usepackage[bookmarks=true]{hyperref}
\usepackage{multicol,amsmath,amssymb,times,xcolor,graphicx,float}
\usepackage{algorithm,algorithmicx,algcompatible,multicol,caption}

\floatname{algorithm}{Algorithm Comparison} 
\newcommand{\algcomment}[1]{%
    \vspace{-\baselineskip}%
    \noindent%
    {\footnotesize #1\par}%
    \vspace{\baselineskip}%
    }

\title{\LARGE \bf Bayesian Particles on Cyclic Graphs}

\author{Ana Pervan$^{1}$ and Todd D. Murphey$^{1}$
\thanks{$^{1}$Department of Mechanical Engineering, Northwestern University, Evanston, Illinois, USA
        {\tt\small anapervan@u.northwestern.edu}
        {\tt\small t-murphey@northwestern.edu}%
}}

\begin{document}

\maketitle

\begin{abstract}

We consider the problem of designing synthetic cells to achieve a complex goal (e.g., mimicking the immune system by seeking invaders) in a complex environment (e.g., the circulatory system), where they might have to change their control policy, communicate with each other, and deal with stochasticity including false positives and negatives---all with minimal capabilities and only a few bits of memory.

We simulate the immune response using cyclic, maze-like environments and use targets at unknown locations to represent invading cells. Using only a few bits of memory, the synthetic cells are programmed to perform a reinforcement learning-type algorithm with which they update their control policy based on randomized encounters with other cells. As the synthetic cells work together to find the target, their interactions as an ensemble function as a physical implementation of a Bayesian update. That is, the particles act as a particle filter.

This result provides formal properties about the behavior of the synthetic cell ensemble that can be used to ensure robustness and safety. This method of simplified reinforcement learning is evaluated in simulations, and applied to an actual model of the human circulatory system.

\end{abstract}

\section{Introduction}
As robot size decreases to the order of a single cell, previously inconceivable applications and abilities emerge. These include monitoring of oil and gas conduits \cite{koman2018}, electrophysiological recordings with neural dust motes \cite{seo2016}, minimally invasive medical procedures \cite{sitti2015applications}, and much more. In this work, we investigate the use of synthetic cells to imitate some of the functionality seen in the immune system. 

The immune system protects the body by recognizing and responding to antigens, which are harmful agents like viruses, bacteria, and toxins \cite{chaplin2010immune}. When white blood cells find a target, they multiply and send signals to other cells to communicate their discovery \cite{newman2018immune}. We show that a group of synthetic cells can imitate this discovery and communication behavior, by collectively executing a type of reinforcement learning that manifests itself as a Bayesian update over the control policy that brings cells to the location of an antigen. 

Synthetic cells are microscopic devices with limited sensing, control, and computational capabilities \cite{janus}. They can contain simple circuits that include minimal sensors and very limited nonvolatile memory---barely a handful of bits \cite{strano}. These devices are around 100$\mu$m in size, rendering classical computation using a CPU impossible. But simple movement, sensory, and memory elements can potentially be combined with a series of physically realizable logical operators to enable a specific task \cite{pervan2018wafr, liu2020} (e.g., a simple reinforcement learning algorithm for finding a target) and communication about how to achieve that task.

Reinforcement learning is centered on finding a suitable action to maximize a reward in a particular situation \cite{sutton1998rl}. Unlike in supervised learning, where an agent is trained using examples of optimal outputs, in reinforcement learning an agent must choose its actions by learning through experience. This is especially important in applications related to medicine and biology, where no two situations are the same. Robots cannot be preprogrammed to perfectly achieve a task in such an environment---we can expect that they must have some element of online learning, and in the context of synthetic cells the question is how they can learn without traditional computation.

A version of reinforcement learning can be executed in a group of synthetic cells that begin with different control policies---indicating \emph{how}, and implicitly where, they should explore---and then communicate with each other that they have or have not been successful in detecting a target. After communicating their success, some synthetic cells will change their control policies to reflect the successes of others in the group. Thus the distribution of synthetic cell control policies reflects the expected location of the target.

\begin{figure}
\centering
\includegraphics[width=0.95\linewidth]{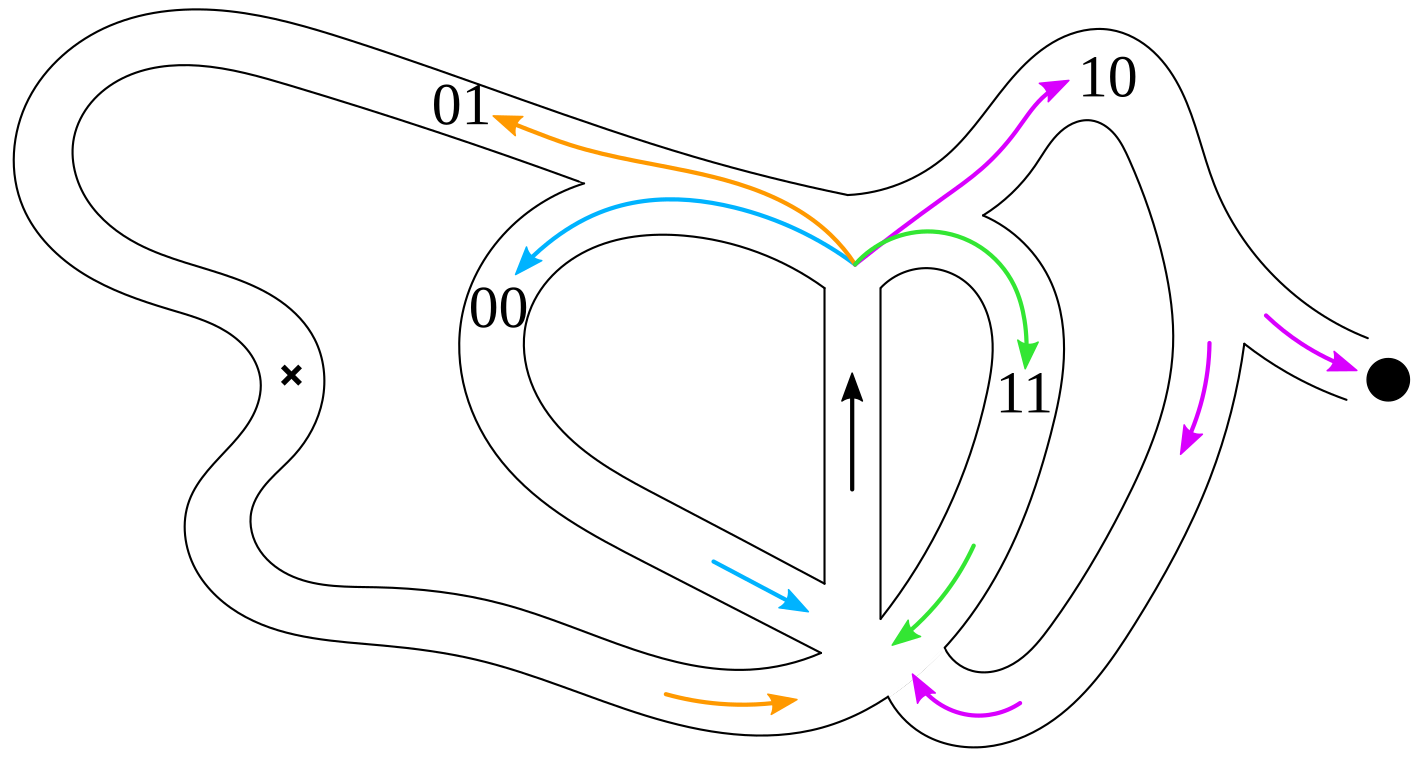}
\caption{\label{fig:complex_maze} A cyclic maze that mimics aspects of the circulatory system. A group of synthetic cells would require at least three bits each to navigate through the four paths and record instances of target detection. One of the paths ($10$) leads to a juncture where cells might get lost and not return.}
\end{figure}

This update is similar to a particle filter, where samples from a distribution are represented by a set of particles, and each particle has a likelihood weight assigned to it that corresponds to the probability of that particle being sampled from the distribution. Particle filters also include a resampling step, to mitigate weight disparity before the weights become too uneven, which closely mirrors the communication step in this synthetic cell implementation, as we discuss in Section \ref{sec:simulations}.

In this paper we show how synthetic cells can use simple, local algorithms and only a few bits of memory to enable global reinforcement learning behavior to refine their belief of a target location. We also show that this implementation of the ensemble of synthetic cells is a suboptimal Bayesian filter, where the control policy of the cells is the decision variable. By constraining synthetic cells to behave as a Bayesian update, the group of cells inherits formal properties in the form of guarantees on asymptotic performance and probabilistically predictable behavior. These properties will help us to reason about robustness and safety in task execution. That is, we are replacing the model of a distributed system with a single Bayesian filter.

\section{Related Work}
\label{sec:relatedwork}
Literature surveys of previous work on nanotechnology and mobile microrobots can be found in \cite{strano} and \cite{sitti2015applications}, respectively. In the discussion of existing challenges associated with designing miniaturized robots for biomedical applications, \cite{sitti2015applications} notes that most robots with dimensions less than $1mm$ use an ``off-board" approach where the devices are externally actuated, sensed, controlled, or powered. In this work, we employ fully autonomous devices that process information and act independently of external drivers and centralized computers.

Research in nanofabrication and synthesis methods have yielded sophisticated synthetic devices, including particles that serve a particular function (e.g., light control for nanoactuation \cite{ding2016}, performing clocked, multistage logic \cite{yao2014}, actuation using external magnetic fields \cite{sitti2009}, \cite{zhang2009}), but not particles possessing autonomous circuitry, logic manipulation, and information storage \cite{koman2018}. Besides the work published in \cite{koman2018, liu2020}, existing micro- or nanoparticles do not autonomously process information when decoupled from their environment \cite{kaewsaneha2015}, \cite{kamaly2016}.  The particles created in Koman and Liu et al. \cite{koman2018, liu2020} are the basis for the synthetic cells proposed in this paper, and we will make the following assumptions based on this work:

\begin{enumerate}
   \item Synthetic cells can guide their own motion either mechanically \cite{guix2014}, by means of elaborate swimming strategies like rotating helical flagella \cite{lenaghan2014}, or (more likely) chemically \cite{moran2017,moran2019}, through use of Pt-Au bimetallic rods \cite{paxton2004}, self-electrophoresis \cite{yuping2007}, \cite{moran2010}, self-diffusiophoresis \cite{howse2007}, or self-thermophoresis \cite{jiang2010}.
   \item Synthetic cells can send and receive communications optically, using integrated LEDs \cite{wu2018} and solid-state or organic light emitting diodes \cite{wu2018,lee2018}.
   \item Synthetic cells can detect a target by using a chemiresistor to recognize the target's specific chemical analyte \cite{koman2018,janus,strano}.
\end{enumerate}

Lastly, and most importantly, \cite{sitti2015applications} discusses that it is mandatory to guarantee the safety and robustness of biomedical microrobots while they are operating inside a human body. If devices are to be employed in medical applications, they must not damage tissues or cause any negative reaction from the body. One of the primary contributions of the work in this paper is constraining synthetic cells to behave as a physical implementation of a Bayesian update, creating a basis for formal properties and guarantees on their behavior. This ensures robustness in their performance, which can be translated to safety guarantees in specific situations and environments.

Bayesian approaches have been applied to reinforcement learning \cite{price2003}, \cite{ramachandran2007}, but often in supervised scenarios where there are experts and learners. These methods also often employ passive observations by the learners, rather than explicit communication from experts to learners. In this paper, we employ reinforcement learning of policy updates in rollouts of executions. We also enforce communication between agents, specifically from successful agents to unsuccessful ones when they encounter each other.

\section{Problem Definition}
\label{sec:problem}

\subsubsection{Environment}
\begin{figure}
\centering
\includegraphics[width=\linewidth]{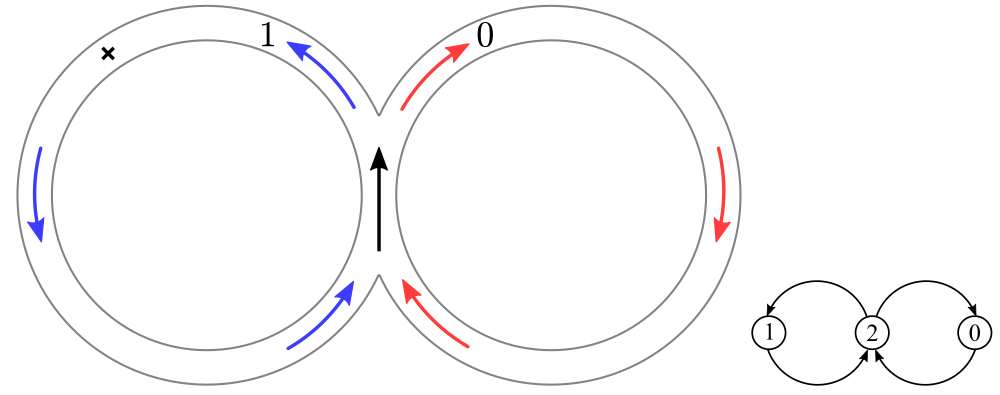}
\caption{\label{fig:2bitmap} Left: An example cyclic maze with only two possible paths. Based on the control policy of each two-bit synthetic cell (a $0$ or a $1$), the cell will follow either the right path or the left path in search of the target \textbf{$\times$}. The cell uses its second bit to encode whether or not it has detected the target. Right: A graphical representation of this synthetic cell environment.}
\end{figure}

Our goal is to mimic the immune response, so we simulate a model of the circulatory system \cite{hardy1982} in Section \ref{sec:simulations}. But in this section, we present a simplified, introductory model. This model consists of a maze, shown in Figure \ref{fig:2bitmap}, with only two possible paths: left or right. The cells will search for a target, located at the black \textbf{$\times$}. 

\subsubsection{Policy Execution}
\begin{figure}
\centering
\includegraphics[width=\linewidth]{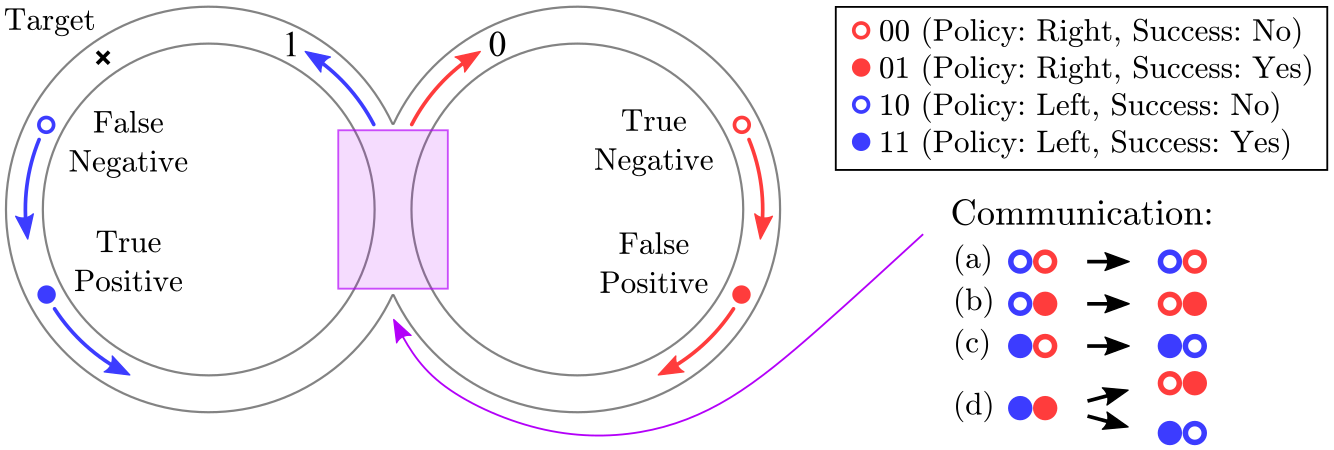}
\caption{\label{fig:env} Each synthetic cell begins with an initial control policy of either a $1$ or a $0$, which causes it to turn either left or right in the maze. If the cell thinks it has found the target, it changes its second bit (its success bit) to a $1$. Due to stochasticity in the environment and the cells, there is the possibility of a false positive or a false negative. Cells might communicate in the middle region, shown in purple. (a) If both cells are unsuccessful they will not communicate any information even if they are within range of each other. (b) and (c) If one cell is successful and one isn't, the successful cell will communicate its policy to the unsuccessful one. (d) If both cells are successful, one (selected randomly) will listen to the other.}
\end{figure}

 For this introductory example, each synthetic cell has only two bits: one for its control policy ($1$ for left or $0$ for right) and one to indicate whether it has found the target ($1$) or not ($0$). Each cell begins with a randomly assigned control policy and loops through the maze. They have a probability of a false positive $p_{fp}$ (detecting the target when it is \emph{not} there) and a probability of a false negative $p_{fn}$ (\emph{not} detecting the target when it is there). If a synthetic cell thinks it has detected the target, it changes its second bit, which we will call its \emph{success bit}, to a $1$. This policy execution is illustrated in Figure \ref{fig:env}.

\subsubsection{Communication}
Using methods of optical information transmission discussed in Section \ref{sec:relatedwork}, synthetic cells are capable of local communication when they are within a certain distance of each other.

When synthetic cells reconvene in the middle of the maze, there is an opportunity for communication. We use a parameter $\rho$ to characterize how many other synthetic cells, on average, each cell will interact with during one loop of the maze (no matter what control policy or success bit either cell has). This parameter $\rho$ is related to the density of synthetic cells in the environment, and how likely they are to pass within communication range of each other. 

If two synthetic cells come into contact, a successful cell (with a $1$ for its success bit) will tell an unsuccessful cell (with a $0$ for its success bit) its ``correct" policy---even if it is successful because of a false positive. If both communicating synthetic cells are successful, one will listen to the other, but which one is the listener is randomly chosen. And if both are unsuccessful they will not tell each other anything. In this way, the success bit also functions as a read/write bit. If it is a $0$, the cell will listen to others (read) and if it is a $1$, the cell will try to broadcast its policy to others (write).  These different scenarios are depicted in Figure \ref{fig:env}, and a visualization of this system is shown in the supplementary video.

\begin{figure}
\centering
\includegraphics[width=\linewidth]{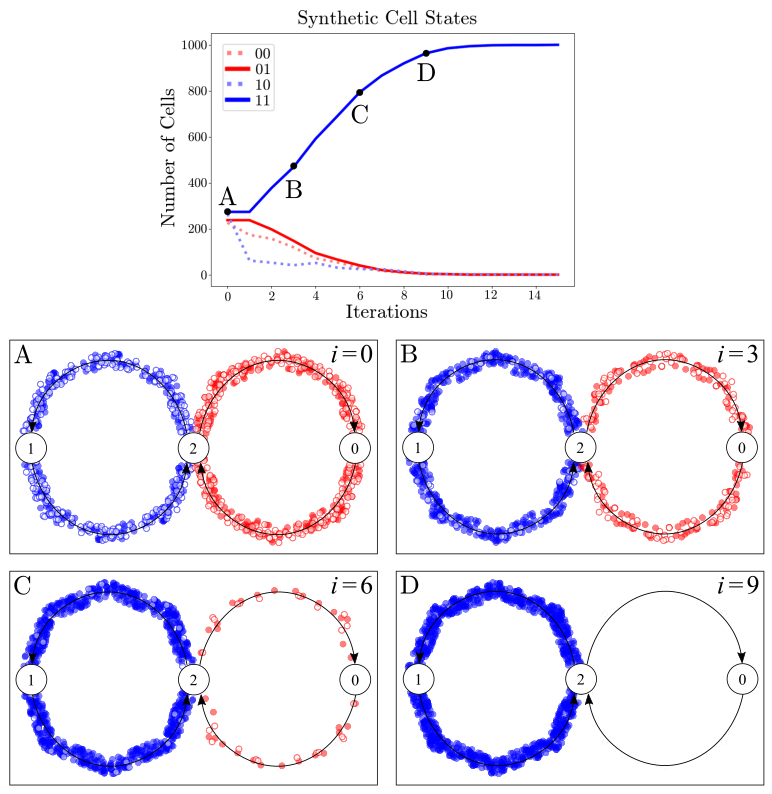}
\caption{\label{fig:2bit} Top: Results of a simulation with $1000$ two-bit synthetic cells for $15$ iterations ($15$ loops around the maze shown in Figure \ref{fig:2bitmap}), with parameters $p_{fp} = 0.2$, $p_{fn} = 0.2$, and $\rho = 1.0$. Bottom: Density plots illustrate the distribution of cells at different time increments. As the cells loop through the maze, they converge to the policy that takes them past the target.}
\end{figure}

When synthetic cells enact their simple algorithm of policy execution, possible target detection, and communication, the cells all end up with the policy that passes the target. In Figure \ref{fig:2bit}, the number of synthetic cells with each state are shown as they loop through the maze multiple times and communicate with each other between loops. By the ninth iterate, every synthetic cell has the policy that takes it past the target. This optimal final result always occurs in the case of this simple maze, as long as $p_{fp}$ and $p_{fn}$ are sufficiently small and $\rho >0$. But with more complicated environments and possibilities (for example, the maze in Figure \ref{fig:complex_maze}) it becomes more difficult to ensure this result. To address this, we increase the number of bits on each cell, to extend their capabilities.

\section{Simulations}
\label{sec:simulations}

We now introduce a more complex example, where each synthetic cell has three bits and the maze has four possible paths. The environment is shown in Figure \ref{fig:complex_maze}, where the possible control policies are: $01$, which takes the synthetic cells past the target; $00$ and $11$, which both loop the cells around the maze; and $10$, which leads the cells down a path that they have probability $p_{lost}$ of never returning from. The goal is for as many cells as possible to end with the $01$ policy, where they will all be heading toward the target.

\setlength{\belowcaptionskip}{-5pt}
\begin{algorithm*}
\caption{}
\begin{multicols}{2}
\begin{algorithmic}
  \STATE \textbf{Particle Filter} 
  \begin{small}
  \STATE \textbf{Prior} distribution is described by $L$ particles and their weights 
  \STATE \hskip2.0em $X_n, w_n$
  \STATE Distribution is sampled, resulting in $L$ \textbf{new particles} 
  \STATE \hskip2.0em $\bar{x}_{n+1}^1, \bar{x}_{n+1}^2, ..., \bar{x}_{n+1}^L$
  \STATE Based on a \textbf{measurement}, weights$^1$ are assigned to each particle
  \STATE \hskip2.0em $w_{n+1}^{\ell} = P(z_{n+1}|\bar{x}_{n+1}^{\ell})$
  \STATE \textbf{Resample} by drawing with replacement $L$ particles from weighted set $\bar{X}_{n+1}$
  \STATE \hskip2.0em 
  \STATE \textbf{Posterior} distribution is described by the resampled particles and their weights
  \STATE \hskip2.0em $X_{n+1}, w_{n+1}$
  \end{small}\\
  
  \STATE \textbf{Synthetic Cell Implementation}
  \begin{small}
  \STATE \textbf{Prior} distribution is described by $M$ cell policies and success bits
  \STATE \hskip2.0em $Y_n, s_n$
  \STATE Cells execute their policies, resulting in $M$ \textbf{new states}
  \STATE \hskip2.0em $\bar{y}_{n+1}^1, \bar{y}_{n+1}^2, ..., \bar{y}_{n+1}^M$
  \STATE Based on its \textbf{success bit}, a cell might broadcast its policy
  \STATE \hskip2.0em $s_{n+1}^m = 0$ or $1$
  \STATE Each cell $\bar{y}_{n+1}^m$ \textbf{communicates} with $\rho$ other cell(s). If any cell has $s_{n+1}^m = 1$, some cell(s) will change their policy.
  \STATE \hskip2.0em This approximates resampling as $\rho \rightarrow M$.
  \STATE \textbf{Posterior} distribution is described by the final synthetic cell policies and success bits
  \STATE \hskip2.0em $Y_{n+1}, s_{n+1}$
  \end{small}
\end{algorithmic}
\end{multicols}
\begin{small}
 \algcomment{$^1$In the case of this synthetic cell example, the measurement $z_{n+1}$ is the number of cells with each policy. The weight (likelihood of measurement $z_{n+1}$ occurring if hypothesis $\bar{x}_{n+1}^{\ell}$ is correct) is calculated by the number of observed particles with the same policy as $\bar{x}_{n+1}^{\ell}$ divided by $L$.}
  \end{small}
\end{algorithm*}

\begin{figure}[t]
\centering
\includegraphics[width=.9\linewidth]{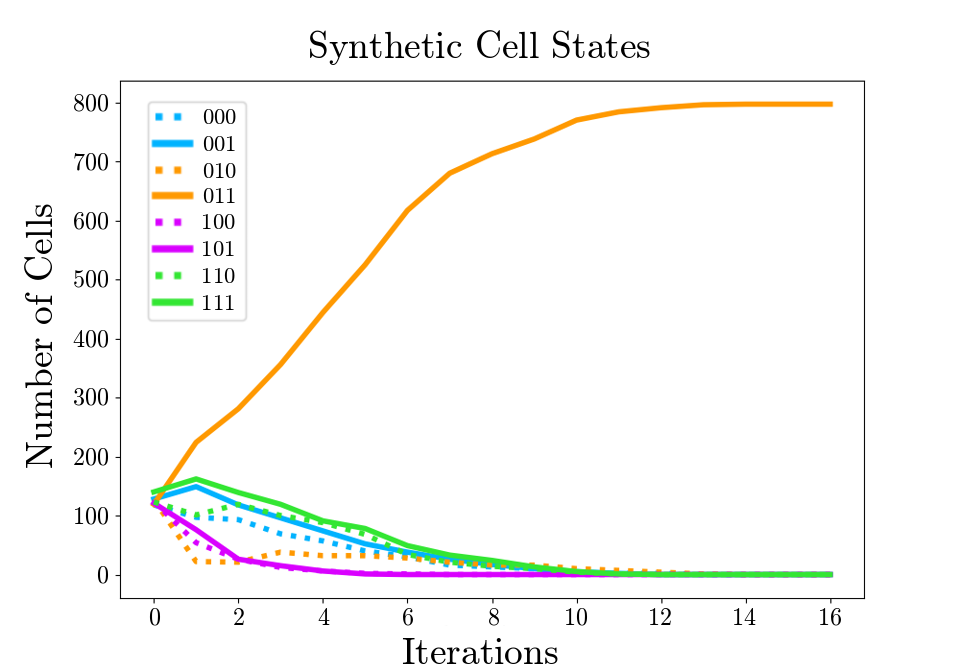}
\caption{\label{fig:3bitstates} Simulated results for $1000$ three-bit synthetic cells executing their policies and communicating in the maze from Figure \ref{fig:complex_maze}, with parameters $p_{fp} = 0.2$, $p_{fn} = 0.2$, $p_{lost} = 0.5$, and $\rho = 1$. Around $800$ of them converge to the correct policy where they will find the target.}
\end{figure}

The possibility of getting lost adds further complexity to the system, because not only is the target not reachable with policy $10$, but some cells with that policy will not return at all. This could be equivalent to different environmental factors in a body, for example an area with enough acidity to damage or destroy synthetic cells. In practice, we predict that there will be many opportunities for synthetic cells to veer off course and get lost, or to get stuck such that they can no longer contribute to the goals of the group. As the magnitude of $p_{lost}$ increases, more cells get lost and fewer are able to reach the target and combat an invading antigen.

Figure \ref{fig:3bitstates} shows results for $1000$ three-bit synthetic cells exploring the environment shown in Figure \ref{fig:complex_maze}. All but the cells that have been lost converge to the correct policy after $12$ iterations.

\subsection{Particle Filter}

A particle filter is a nonparametric implementation of a Bayes filter that represents a distribution using a set of random samples drawn from that distribution \cite{bishop}, \cite{thrun}. In a particle filter, the samples from the distribution are called \textit{particles}. We denote these samples $X_n:=x_{n}^{1},x_{n}^{2},...,x_{n}^{L}$. Each particle $x_n^{\ell}$ is a hypothesis of the true world state at time $n$---in our example, each particle would be a hypothesis of the policy that leads to the target.

\begin{figure*}
\centering
\includegraphics[width=\textwidth]{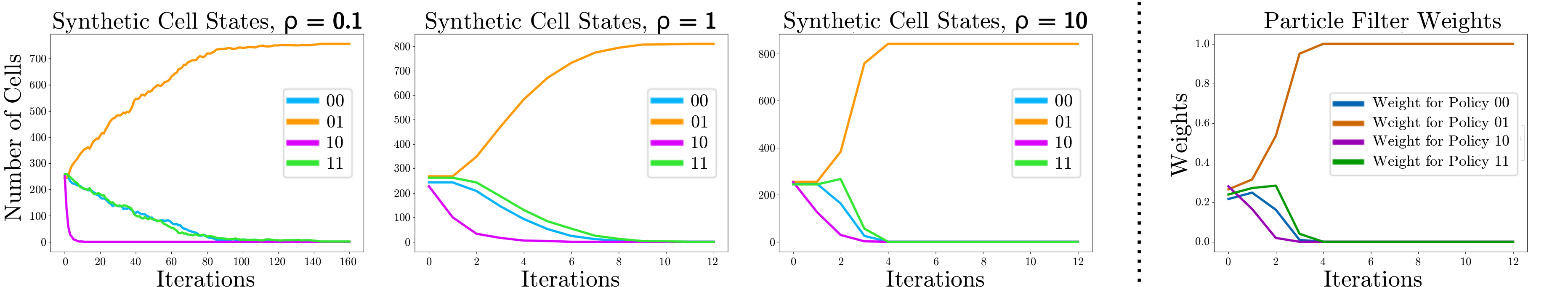}
\caption{\label{fig:diff_rho} Synthetic cell executions for different values of $\rho$ for $1000$ synthetic cells and $p_{fp} = 0.1$, $p_{fn }= 0.1$, $p_{lost} = 0.5$. The rightmost panel shows a particle filter implementation for this system, where the measurements are the current states of the synthetic cells. As $\rho$ increases, it approximates the particle filter. Note that the third plot, where $\rho=10$, is nearly identical to the plot of the particle filter weights.}
\end{figure*}

The most basic variant of a particle filter algorithm begins with the particle set $X_{n}$ and weights $w_{n}$, which together represent a prior distribution. This distribution is sampled, resulting in $L$ particles $\bar{x}_{n+1}^1, ..., \bar{x}_{n+1}^L$. The bar indicates that these samples are taken before the measurement has been incorporated. Next, a measurement $z_{n+1}$ is obtained, and it is used to calculate new weights $w_{n+1}^{\ell}$ for each particle. The weight is the probability of the measurement given each particle, $w_{n+1}^{\ell} = P(z_{n+1}|\bar{x}_{n+1}^{\ell})$. Lastly, the particle filter resamples the distribution by drawing with replacement $L$ particles from the weighted set $\bar{X}_{n+1}$, where the probability of drawing each particle is given by its weight. The resampled particles $X_{n+1}=x_{n+1}^1, ..., x_{n+1}^L$, along with the weights $w_{n+1}$, represent the posterior distribution---an updated estimate of which policy leads to the target. Note that in the case of the synthetic cell ensemble, the particle filter is estimating discrete states: each particle can only take one of four different values. There are much more than four particles, so many particles will hypothesize that the target is at the same state.

For the synthetic cell implementation, we begin with random policies (and random success bits) on all of the synthetic cells, similar to starting with a uniformly distributed prior distribution. This distribution of $M$ synthetic cells has discrete states $Y_{n} := y_n^1, y_n^2, ..., y_n^M$. The cells execute their policies and some return with a success bit, resulting in new cell states $\bar{y}_{n+1}^1, ..., \bar{y}_{n+1}^M$.  The value of the success bit $s_{n+1}^m$ of each cell $\bar{y}_{n+1}^m$ is a binary implementation of the weight $w_{n+1}^m$ in a particle filter. Instead of calculating a conditional probability so that the weights are \textit{between} $0$ and $1$, the weights \textit{are} either $0$ or $1$ (before being normalized by the total number of successful cells). A cell $\bar{y}_{n+1}^m$ with a $0$ success bit will not communicate its policy to any other cells, meaning, in particle filter terms, that it will not be sampled from---so its weight is effectively $w_{n+1}^m=0$. 

For example, consider a situation where there are $M$ synthetic cells (and $M_{001}$ synthetic cells with policy $00$ and a $1$ for a success bit, $M_{110}$ cells with policy $11$ and a $0$ for a success bit, etc.) and $\rho=1$, meaning that each cell will communicate with one other cell during each loop around the maze. Cell $\bar{y}_{n+1}^m$ has a $\frac{\rho}{M}$ probability of communicating with any other cell $\bar{y}_{n+1}^i$, where $1 \leq i \leq M$, during a given cycle. Cell $\bar{y}_{n+1}^m$'s probability of sampling a cell with policy $00$ is $\frac{M_{001}}{M}$, its chance of sampling a cell with policy $01$ is $\frac{M_{011}}{M}$, and so on. It also has a chance of staying the same, anytime it communicates with a cell with a $0$ success bit, which occurs with probability $\frac{M_{000}+M_{010}+M_{100}+M_{110}}{M}$. 

This is the main difference between the particle filter algorithm and the synthetic cell implementation: the synthetic cells have some probability of \textit{not} resampling, and just staying the same---unlike particles in a particle filter which are all resampled, every iteration. This difference is demonstrated in the Algorithm Comparison box, above. If each cell \textit{always} communicated with a random successful cell, its behavior would be the same as that of a particle filter. This is illustrated in Figure \ref{fig:diff_rho}. The far right panel of Figure \ref{fig:diff_rho} shows a particle filter applied to the synthetic cell system. There are $L=1000$ particles being randomly sampled from the synthetic cell distribution (which is also comprised of $M=1000$ cells), and the weights are being updated based on observations of synthetic cell policies.  As $\rho$ increases, the amount of resampling increases, and the synthetic cell behavior is guaranteed to converge to the particle filter behavior. The physical execution of the group of synthetic cells approximates the particle filter algorithm. 

Similarly, a particle filter approximates a Bayes filter. The approximation error of a particle filter approaches zero as the number of particles goes to infinity---\emph{the error depends on the number of particles, not on the resampling}. In fact, an alternative version of a particle filter does not resample at all \cite{thrun}. Since the asymptotic guarantee on a particle filter approximating a Bayes filter does not depend on resampling, it consequently holds for synthetic cells as well. Therefore, since the synthetic cell implementation approximates a particle filter, and a particle filter approximates a Bayesian update, we can conclude that a synthetic cell system using this algorithm approximates a Bayesian update.

This result, which is illustrated in Figure \ref{fig:diff_rho}, guarantees convergence properties for how synthetic cells will probabilistically behave. These guarantees are valuable because they can be used to reliably predict how synthetic cells will perform in new scenarios, and we can be certain of robustness and safety requirements for physical experiments. 

\subsection{Model of the Human Circulatory System}

\begin{figure}[h]
\centering
\includegraphics[width=.9\linewidth]{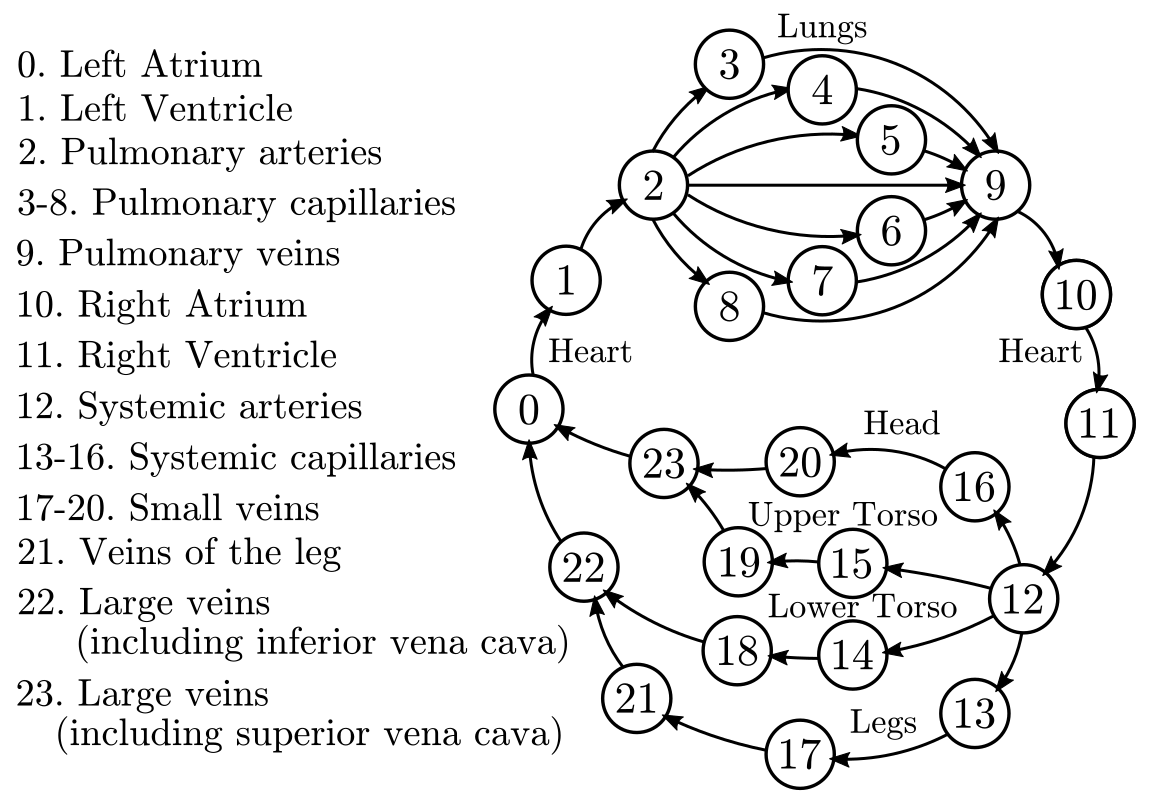}
\caption{\label{fig:circulatory} Adapted from Figure 1, Figure 2, Figure 3, and Table 2 in Hardy~et~al.~\cite{hardy1982}. Each number represents a chamber, as described in the legend. The connections between chambers are inputs and outputs illustrating blood flow.} 
\end{figure}

\begin{figure}
\centering
\includegraphics[width=0.65\linewidth]{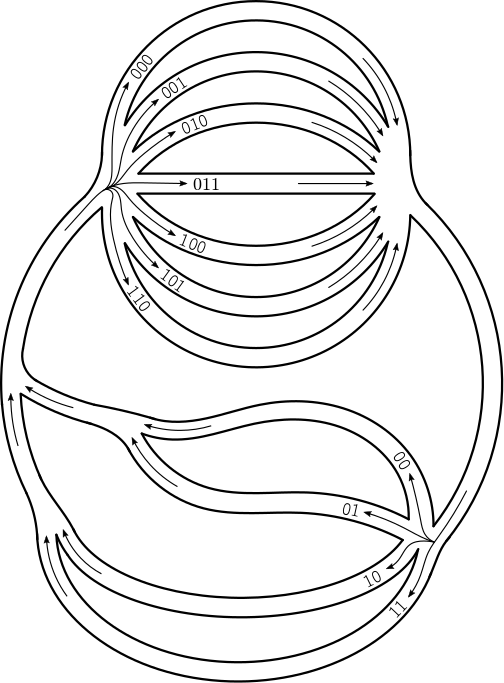}
\caption{\label{fig:circ_maze} A maze, similar to those in Figures \ref{fig:complex_maze} and \ref{fig:2bitmap}, based on the inputs and outputs of chambers in the circulatory system, shown in Figure \ref{fig:circulatory}. Bit assignments for each path are also shown, to illustrate the $5$ bit policies that describe each of the $28$ possible paths through the system.} 
\end{figure}

Many models of the human cardiovascular system exist, including a $36$ vessel body tree \cite{mirzaee2008}, a lumped parameter model \cite{kim2013}, and a mathematical model featuring both linear and nonlinear constitutive relations \cite{conlon2006}. In this paper, we use the model from Hardy~et~al.~\cite{hardy1982}, which clearly defines the $24$ different chambers in the circulatory system, as well as the connections going into and out of each one. This model is shown in Figure \ref{fig:circulatory}, where each number represents a chamber, as described in the legend, and the connections depict inputs and outputs for blood flow. Figure \ref{fig:circ_maze} shows how the graphical representation of the circulatory system can be illustrated as the same type of maze that was shown in Figures \ref{fig:complex_maze} and \ref{fig:2bitmap}.

\setlength{\belowcaptionskip}{0pt}

To navigate in this environment, synthetic cells require seven bits. This comes from the seven way intersection at node $2$, where cells need at least three bits to choose a path; the four way intersection at node $12$, where they each need two more bits to decide on a path; and two more bits to use as success bits (the reason \emph{two} success bits are required will be discussed in the next section). The policy bit organization is shown in Figure \ref{fig:circ_maze}. 

\setlength{\belowcaptionskip}{-10pt}

\begin{figure}[t]
\centering
\includegraphics[width=\linewidth]{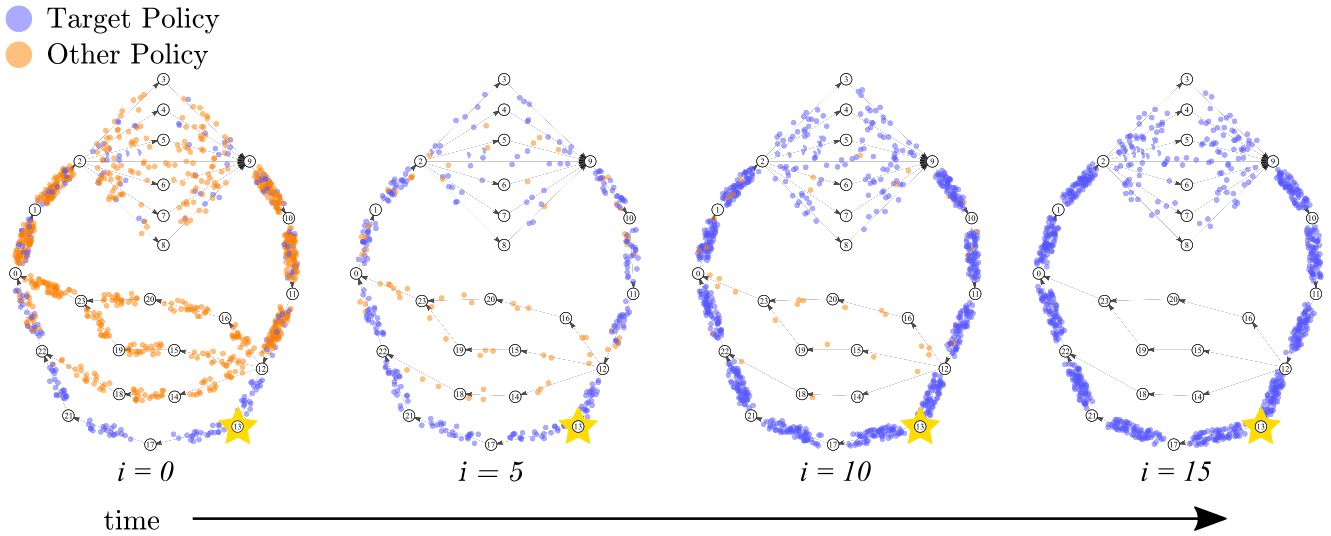}
\caption{\label{fig:7bit} Density plots illustrating the distribution of $1000$ seven-bit synthetic cells executing their policies in the maze from Figure \ref{fig:circ_maze}, with parameters $p_{fp} = 0.1$, $p_{fn} = 0.1$, and $\rho = 1.0$. The target is shown by a yellow star and the cells with policies that pass by it are shown in blue.} 
\end{figure}

We simulated synthetic cell executions in this scenario, where the desired target was in the leg, specifically node $13$. In this circulatory system model, there are many different policies that will lead to finding the target. It doesn't matter how the cells pass through the pulmonary system (states $2-9$ in Figure \ref{fig:circulatory} or the top part of the maze in Figure \ref{fig:circ_maze}), as long as they reach the leg in the end. The results of this simulation are shown in Figure \ref{fig:7bit}.

In the previous example, shown in Figures \ref{fig:complex_maze} and \ref{fig:3bitstates}, we simulated a probability of getting lost along one of the paths. This was to acknowledge that in practice, unexpected events can happen where some synthetic cells will get lost, stuck, destroyed, or otherwise do not contribute to the group's estimate of the target location. We recognize that this can happen no matter where the cell is, so in this example we implemented a small $p_{lost}$ on every execution of every synthetic cell. All of the cells, besides the ones that have been lost to the environment due to $p_{lost}$, converge to the correct policy by about the twenty sixth iteration. A visualization of this system is shown in the supplementary video.

\subsection{Multiple and Moving Targets}
\label{sec:m&m}

In earlier sections, our algorithm was shown to enable all simulated synthetic cells to converge to a policy that passed by a single target. But there will not always be single, stationary targets in the immune system. In this section, we use the same algorithm to show that multiple and moving targets can be found and communicated, without any prior knowledge of the number or movements of the targets.

If two targets in the same environment are in series\footnote{Here, series and parallel mean the same things as they do in circuitry: if two nodes are in series a cell can flow through both of them (e.g., node 3 and node 13, in Figure \ref{fig:circulatory}), and if they are in parallel a cell cannot (e.g., node 13 and node 20).}, the cell needs to be able to distinguish between them. One synthetic cell could be passing by a target in node 3, while another cell is passing by a target in node 13. If they both think they have the correct policy, they might never learn to pass by both targets. If they have separate success bits for each intersection, they ensure that they are reaching as many targets as possible. This conclusion can be expressed in the following equation for the number of bits, $B$, required for any cyclic graph.

\begin{equation}
    \label{eq:bit}
    B = \sum_{i=1}^{I} ceil(log_2(P_i))+I
\end{equation}

In Eq.~\ref{eq:bit}, $B$ is the number of bits required to navigate the graph, $I$ is the number of intersections, or diverging nodes (nodes that have multiple edges leaving them), and $P_i$ is the number of edges leaving each intersection. The ceiling function $ceil$ rounds up to the nearest integer, as we only consider entire bits.

The circulatory system shown in Figure \ref{fig:circulatory} has $I=2$ intersections, at nodes $2$ and $12$, which have $7$ and $4$ outgoing edges, respectively, and therefore $B=7$ bits are required to solve the graph.

\begin{figure}
\centering
\includegraphics[width=\linewidth]{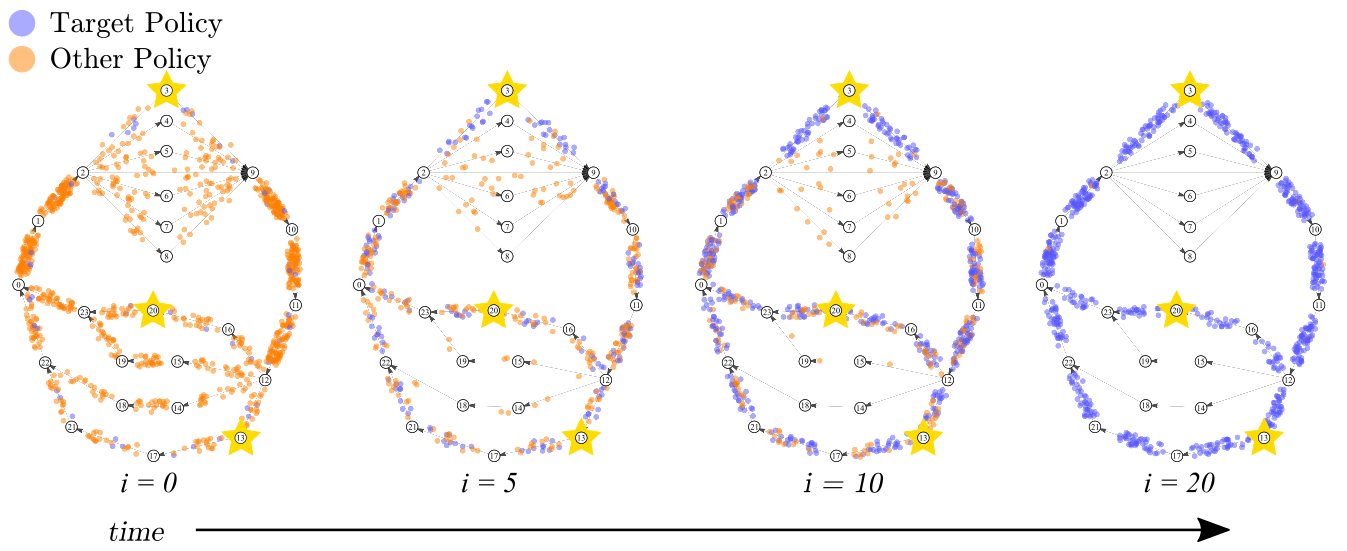}
\caption{\label{fig:multiple} Density plots illustrating the distribution of $1000$ seven-bit synthetic cells executing their policies in the maze from Figure \ref{fig:circ_maze}, with parameters $p_{fp} = 0.1$, $p_{fn} = 0.1$, and $\rho = 1.0$. The targets are at nodes $3$, $13$, and $20$, shown by yellow stars.} 
\end{figure}

Figure \ref{fig:multiple} shows simulated results for $1000$ synthetic cells navigating through an environment that has multiple targets: at nodes $3$, $13$ , and $20$. Within twenty cycles, the cells learn policies to pass by as many targets as possible (in this environment, each cell can pass by a maximum of two targets as they have two decision points throughout the graph).

Next, we investigate moving targets. Once all of the synthetic cells have converged upon a target's location, what if it moves somewhere else? To find it again, the cells will have to rely on a small amount of random exploration and decaying memory. We model the random exploration as a very small chance of one of a cell's bits flipping at any moment (this could also be considered a cell making a mistake). Decaying memory enables synthetic cells' success bits to turn off (back to $0$) after some amount of time has passed since they last detected a target. We know from \cite{koman2018, liu2020} that this is physically feasible, given variable chemical decay rates and reactions that act similarly to capacitors with a decaying charge. 

\begin{figure}[h]
\centering
\includegraphics[width=\linewidth]{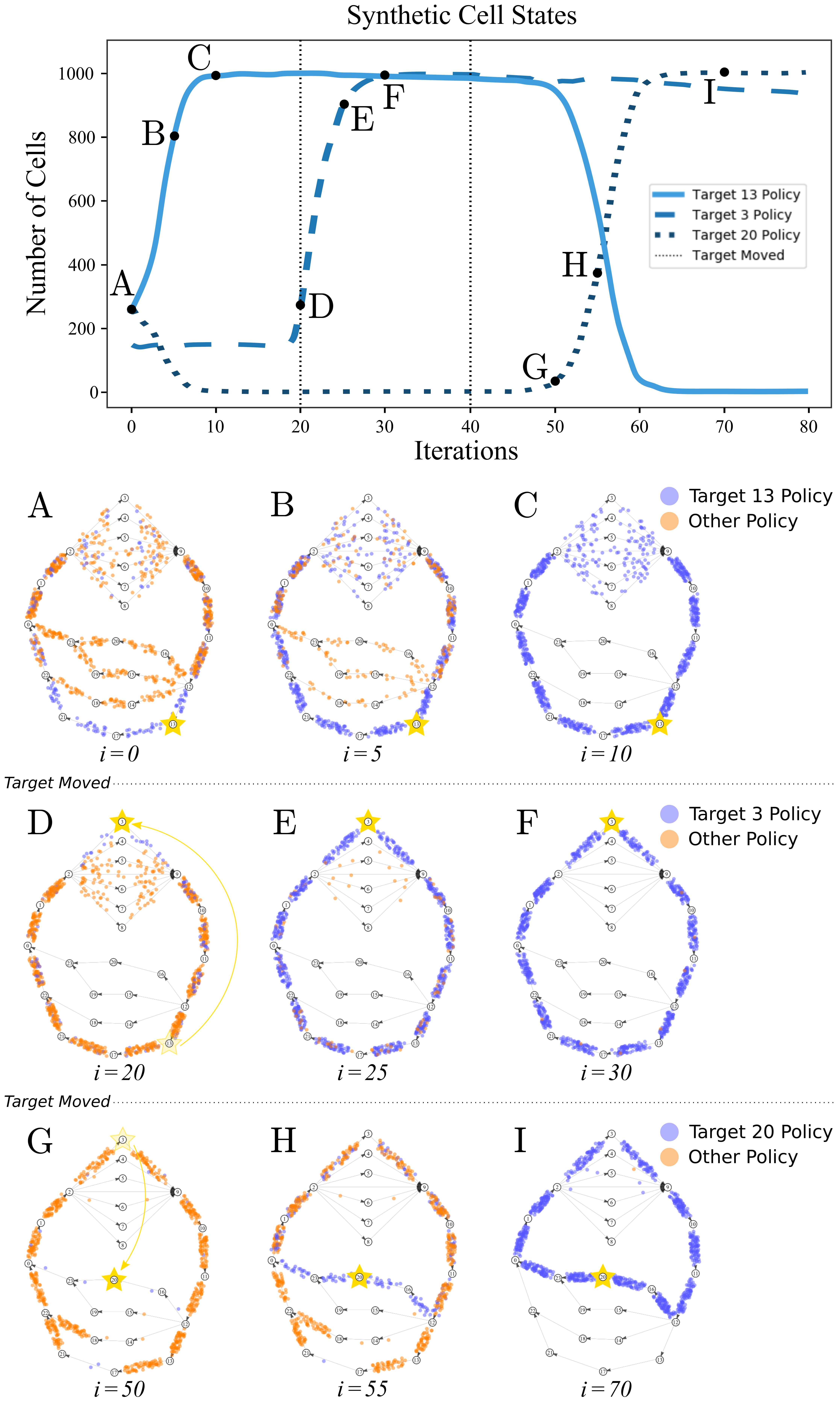}
\caption{\label{fig:moving} Simulated results for 1000 seven-bit synthetic cells executing their policies in the maze from Figure \ref{fig:circ_maze}, , with parameters $p_{fp} = 0.1$, $p_{fn} = 0.1$, and $\rho = 1.0$, as they search for a target that moves from node $13$ to node $3$ to node $20$. At nine instances in time (indicated by letters A-I), a density plot is shown to illustrate the distribution of the cells.} 
\end{figure}

Figure \ref{fig:moving} shows simulated results for $1000$ synthetic cells navigating through an environment and learning the policies to keep finding the new location of a target which moves from node $13$ to node $3$, and finally to node $20$.

\section{Conclusions}
\label{sec:conclusion}

This work demonstrated a novel algorithm for reinforcement learning behavior, in the form of policy updates based on observations, in synthetic cell ensembles. Each synthetic cell only has a few bits of memory and very simple communication abilities. Despite this, we show that the cells can use local algorithms to refine their global belief of how to reach a target location---reflected in the distribution of the control policies of each cell.

This was applied to a model of the human cardiovascular system, where the group of cells was able to converge to the correct policy (apart from those lost to simulated environmental factors), using only seven bits. These same particles were also able to detect and navigate toward multiple targets as well as find and follow a moving target. The result that only seven bits are necessary to function in this model of the circulatory system demonstrates that even with very limited computation synthetic cells are a capable system in an environment as complex as the human body.

We showed that the synthetic cell implementation approximates a particle filter, and that the only difference between the two methods is the execution of the resampling step. Since the asymptotic guarantee on a particle filter approximating a Bayes filter depends on the number of particles, and not the resampling, we concluded that the synthetic cell system is a suboptimal Bayesian filter. This result constitutes what might be the first decision theoretic model of the immune system, and provides formal properties for the behavior of this type of synthetic cell ensemble that can be applied in future work with different tasks, environments, and decision variables. 

In the future, we will pursue different elements of this work, including encoding the need to explore (e.g., if the correct policy is not known to any cells in the ensemble, or if the target has been successfully destroyed). To do this we will borrow methods from \cite{otte2018}, which demonstrates effective algorithms for path planning of long excursions that agents may not return from.  We also intend to implement these results experimentally in the near future.

\addtolength{\textheight}{-12cm}   

\section*{Acknowledgments}
We greatly appreciate Albert Tianxiang Liu and Michael Strano for their feedback on this manuscript.

\bibliographystyle{IEEEtran}
\bibliography{references}

\end{document}